%% file: main.tex
\documentclass[10pt,twocolumn,letterpaper]{article}

\usepackage[accsupp]{axessibility} 
\usepackage{multirow}
\usepackage{csquotes}
\usepackage{xcolor}
\usepackage{float}
\usepackage{subcaption}

\usepackage[applications]{wacv}
\definecolor{wacvblue}{rgb}{0.21,0.49,0.74}
\usepackage[pagebackref,breaklinks,colorlinks,allcolors=wacvblue]{hyperref}

\usepackage[table]{xcolor}
\usepackage{colortbl}
\newcommand{\best}[1]{\cellcolor[gray]{0.85}\textbf{#1}}


\def\wacvPaperID{712}
\def\confName{WACV}
\def\confYear{2027}

\title{Multimodal-Conditioned Latent Diffusion Models for\\Spatiotemporal Brain MRI Forecasting}

\author{
Antonio Scardace$^{1}$ \quad Silvia Ingala$^{2,3,4}$ \quad Sebastiano Battiato$^{1}$ \quad Daniele Ravì$^{5}$\\[2pt]
{\small $^{1}$University of Catania \quad $^{2}$Copenhagen University Hospital -- Rigshospitalet}\\
{\small $^{3}$Copenhagen University Hospital -- Herlev and Gentofte \quad $^{4}$Cerebriu A/S \quad $^{5}$University of Messina}\\[2pt]
{\tt\small antonio.scardace@phd.unict.it, silvia.ingala.01@regionh.dk}\\
{\tt\small sebastiano.battiato@unict.it, daniele.ravi@unime.it}
}

\begin{document}
\maketitle

%% ---

\begin{abstract}

Spatiotemporal forecasting of 3D anatomical changes is a critical generative vision challenge for modeling Alzheimer’s Disease (AD) progression. While brain MRI forecasting from single-modality data has advanced, other key complementary multimodal biomarkers remain underutilized. In this work, we propose the \textbf{M}ultimodal-\textbf{C}onditioned \textbf{L}atent \textbf{D}iffusion \textbf{M}odel (\textbf{MC-LDM}), a novel framework for patient-specific brain MRI synthesis. MC-LDM leverages multimodal biomarkers---MRI, tau-PET, Clinical Dementia Rating (CDR), and APOE $\varepsilon4$---during training, but relies only on baseline MRI, APOE $\varepsilon4$, and patient metadata at inference, reflecting realistic clinical availability. Our contributions are threefold: (1) a Latent Diffusion Model (LDM) fine-tuned via Low-Rank Adaptation (LoRA) for data-efficient learning; (2) a multimodal conditioning that integrates molecular and clinical signals to model heterogeneous disease trajectories; and (3) auxiliary predictors estimating unobserved multimodal biomarkers from MRI features and patient covariates to condition the diffusion model, removing the need for these assessments at inference. Trained on 1,355 visits from 852 subjects in the Alzheimer’s Disease Neuroimaging Initiative (ADNI), MC-LDM reduces average regional volumetric MAE by 18\% over the second-best baseline on an independent test set of 664 visits. To our knowledge, MC-LDM represents the first diffusion-based framework to leverage multimodal priors for MRI forecasting. This conceptual advancement enables patient-specific trajectory generation that captures aspects of AD heterogeneity not observable from structural anatomy alone, bridging the gap between rich multimodal supervision and the strict constraints of routine clinical deployment. Code is available at: \url{https://github.com/antonioscardace/MC-LDM}.

\end{abstract}

%% ---

\section{Introduction}

Forecasting 3D anatomical changes is a challenging spatiotemporal problem in generative computer vision, particularly for modeling progressive pathologies such as Alzheimer’s Disease (AD). As an age-related neurodegenerative disorder affecting over 57 million people worldwide, AD is projected to nearly triple in prevalence by 2050~\cite{ADstats}. It typically manifests in older adults through insidious cognitive decline, memory impairment, and behavioral alterations that severely disrupt daily life activities and ultimately lead to a complete loss of independence~\cite{Ad2018}. As a result, its clinical management critically depends on early detection and accurate monitoring of structural and pathological brain changes over time.

In clinical practice, capturing the temporal evolution of AD may require a multimodal assessment involving complementary diagnostic tests to evaluate biomarkers associated with three major pathological processes: amyloid-$\beta$ deposition (A), pathological tau (T), and neurodegeneration (N). Within this ATN framework~\cite{Ad2016,Ad2018}, each biomarker category is classified as either positive or negative.

Among genetic factors, the APOE $\varepsilon4$ allele is the strongest known risk factor for AD, promoting early amyloid-$\beta$ accumulation through impaired peptide clearance and reducing neuronal injury response~\cite{Mahley2012,liu2013apolipoprotein}. While this genetic vulnerability can be readily assessed via blood or saliva sampling~\cite{rajic2025}, its downstream pathological consequence---the accumulation of amyloid deposits in the brain---can be directly visualized using amyloid Positron Emission Tomography (amyloid-PET). In turn, amyloid pathology is thought to drive subsequent tau aggregation, which can be measured via tau-PET and is strongly associated with cognitive decline~\cite{mattsson201918f}.

Despite their strong biological relevance, PET scans remain limited in routine clinical use due to high cost and exposure to radioactive tracers, restricting repeated longitudinal assessments. As AD progresses, these molecular abnormalities, such as amyloid and tau accumulation, contribute to neurodegeneration, ultimately manifesting as structural brain changes. Magnetic Resonance Imaging (MRI) thus remains central to AD evaluation: it enables the assessment of macrostructural biomarkers such as hippocampal atrophy and ventricular enlargement~\cite{frisoni2010clinical}, and is well suited for routine monitoring thanks to its low-cost and non-invasive nature. Finally, clinical measures such as the Clinical Dementia Rating (CDR)~\cite{morris1993} relate these biological and structural changes to functional impairment in daily life.

To support early diagnosis, Disease Progression Modeling (DPM) provides a temporal framework for integrating heterogeneous biomarkers into coherent disease trajectories. Because AD progresses non-linearly over decades with marked inter-patient variability, forecasting individual trajectories is crucial not only for personalized care planning, but also for clinical trial design---enabling the selection of participants expected to exhibit measurable endpoint progression within the trial window to maximize treatment effect size~\cite{Maheux2023}. Methodologically, DPM is commonly categorized into population-level approaches, which estimate average progression patterns across subjects, and individual-level models, which aim to provide personalized forecasts from baseline clinical and imaging data.

Despite recent advances, current state-of-the-art methods still face three critical bottlenecks:

\begin{enumerate}
    \item Many approaches require thousands of aligned longitudinal scans for stable convergence, given the models’ millions of trainable parameters. This issue is further exacerbated when multimodal aligned inputs are introduced, as such datasets are also inherently scarce, increasing the risk of overfitting.
    \item Existing methods typically rely solely on MRI and basic demographic variables, overlooking complementary molecular, cognitive, and genetic information. This limited representation restricts the models’ ability to capture the pronounced heterogeneity of AD progression.
    \item Even when multimodal data are available, practical barriers in clinical settings remain, as key modalities such as tau-PET are rarely acquired longitudinally, leaving MRI as the most accessible imaging tool. Consequently, models that depend on rich multimodal inputs would face significant obstacles to routine clinical deployment.
\end{enumerate}

To address these challenges, we introduce MC-LDM, a multimodal-conditioned latent diffusion framework for individual-level AD progression forecasting from 3D brain MRI. To overcome limitation (1), we explicitly condition the latent diffusion process on molecular, genetic, and clinical signals using Low-Rank Adaptation (LoRA)~\cite{LoRA2022} for parameter-efficient training, mitigating overfitting while guiding MRI synthesis in small multimodal cohorts. To address limitation (2), MC-LDM incorporates multimodal priors into the generative process, allowing more accurate modeling of progression heterogeneity. To address limitation (3), auxiliary models predict regional tau uptake and CDR scores from MRI-derived volumetric features and patient metadata, enabling multimodal supervision during training without requiring extensive clinical assessments at inference. This strategy is motivated by prior evidence that complementary biomarkers from other modalities, such as PET, are reflected in MRI-derived patterns~\cite{Stonnington2010,Sargood2026,Karlsson2026,Scardace2026}, enabling effective cross-modal supervision during training.

In our design, MC-LDM supports both cross-sectional and longitudinal synthesis. To the best of our knowledge, it represents the first generative framework to fully integrate multimodal AD biomarkers into 3D MRI forecasting while remaining clinically practical. Beyond improved forecasting performance, the core novelty of this work lies in advancing DPM towards a practical multimodal paradigm. By adopting an asymmetric conditioning strategy, our framework effectively bridges the gap between rich multimodal supervision during training and the strict missing-modality constraints of real-world clinical deployment.

%% ---

\section{Related Work}

In this section, we review related work on \textbf{(i)} population-level disease progression modeling, and \textbf{(ii)} individual-level disease progression modeling approaches.

\subsection{Population-Level DPM}

Population-level disease progression models aim to learn shared trajectories of disease evolution from longitudinal cohorts, often using statistical formulations such as mixed-effects models~\cite{schiratti2017} to capture inter-subject variability around a common progression pattern.

Recent works have combined these statistical formulations with deep generative models to capture high-dimensional clinical data. For example, \cite{sauty2022} and \cite{chadebec2022} model continuous disease trajectories in low-dimensional Euclidean and Riemannian latent spaces, respectively, while incorporating shared clinical characteristics. However, these methods tend to represent disease progression through shared trajectories within the population, potentially obscuring clinically relevant heterogeneity. To address this issue, recent approaches~\cite{viani2025} have introduced subtype-aware formulations that capture multiple progression patterns.

\subsection{Individual-Level DPM}

Individual-level progression models adopt a flexible formulation, directly predicting future disease states from subject-specific baseline observations. By modeling individual trajectories, they can naturally use chronological age as a temporal reference for personalized forecasts. This contrasts with population-level methodologies, which typically require the estimation of a latent biological timeline to align subjects at different stages of the disease cascade.

Compared to population-level methods, this flexibility often comes at the cost of reduced interpretability, as personalized predictions are less constrained by an explicit shared disease model. Early approaches primarily relied on scalar clinical scores and biomarkers~\cite{oxtoby2017imaging,young2024data}, providing limited spatial interpretability of pathology~\cite{trajectoriesTau2021,eshaghi2021identifying}. More recent spatiotemporal models operate directly on high-dimensional imaging data, enabling voxel-wise localization of structural changes and richer representations of disease progression~\cite{DaniNet2019}. However, a persistent challenge in these latent-space individual architectures is their inability to formally isolate the mathematical specificity of each biomarker in driving the personalized trajectory, often obscuring the underlying phenotypical subgroups.

To generate high-fidelity forecasts, deep generative models have recently gained traction for longitudinal MRI forecasting. Early diffusion-based approaches, such as DDM~\cite{kim2022}, were developed for deformable registration to model anatomical changes between a baseline and a target scan and predict intermediate progression. Because the target scan is required, however, they cannot forecast an unobserved future state. More recent approaches instead aim to predict future anatomical states from baseline information, without requiring the target scan as input. For example, CounterSynth~\cite{CounterSynth2023} employs a GAN-based counterfactual framework. While effective at generating realistic counterfactuals, it can suffer from training instability and limited temporal consistency. Recent works have therefore shifted toward latent diffusion. BrLP~\cite{BrLP2025} leverages temporal information and clinical priors for high-fidelity synthesis, while TADM-3D~\cite{Tadm3d2026} predicts voxel-wise intensity residuals, conditioning forecasts on arbitrary time intervals. MambaControl~\cite{yang2025} guides the diffusion process by explicitly modeling connections between brain regions and the patient’s long-term history. Finally, $\Delta$-LFM~\cite{chen2026} models patient-specific disease progression from longitudinal image data by conceptualizing progression as a continuous velocity field in latent space, enabling individualized temporal evolution over future time points. Despite these advances, current state-of-the-art architectures remain primarily unimodal, relying largely on structural anatomy and basic demographics while overlooking the complementary prognostic value of molecular, genetic, and cognitive biomarkers.

%% ---

\section{Preliminaries}

In this section, we present the core concepts required to construct MC-LDM. We first establish the clinical rationale for targeting structural MRI, before detailing the latent diffusion architecture and the parameter-efficient conditioning mechanisms necessary to integrate multimodal priors.

\subsection{Structural MRI as the Forecasting Target}

A key design choice of our framework is to forecast structural MRI rather than alternative modalities. While techniques such as tau-PET are highly informative in the early and pre-symptomatic stages of AD, they present important physical, economic, and practical limitations~\cite{Moses2011}. More generally, several non-structural modalities, including PET, CSF biomarkers, and functional imaging, are either invasive, costly, less widely available---a key requirement for training our model---or difficult to acquire longitudinally at scale. These limitations directly affect their suitability as prediction targets in real-world clinical settings and constrain their use for training data collection.

In particular, MRI provides high-resolution ($\approx 1$ mm) imaging of brain anatomy, enabling direct assessment of macroscopic neurodegeneration and atrophy, which are closely linked to cognitive decline. MRI also serves as a reference modality for interpreting and correcting PET scans, whose clinical resolution ($\approx 4$–5 mm) is substantially lower, through Partial Volume Correction (PVC)~\cite{PetPvc2016}, further highlighting its central role in multimodal analysis.

\begin{figure*}[t]
    \centering
    \includegraphics[width=0.97\textwidth]{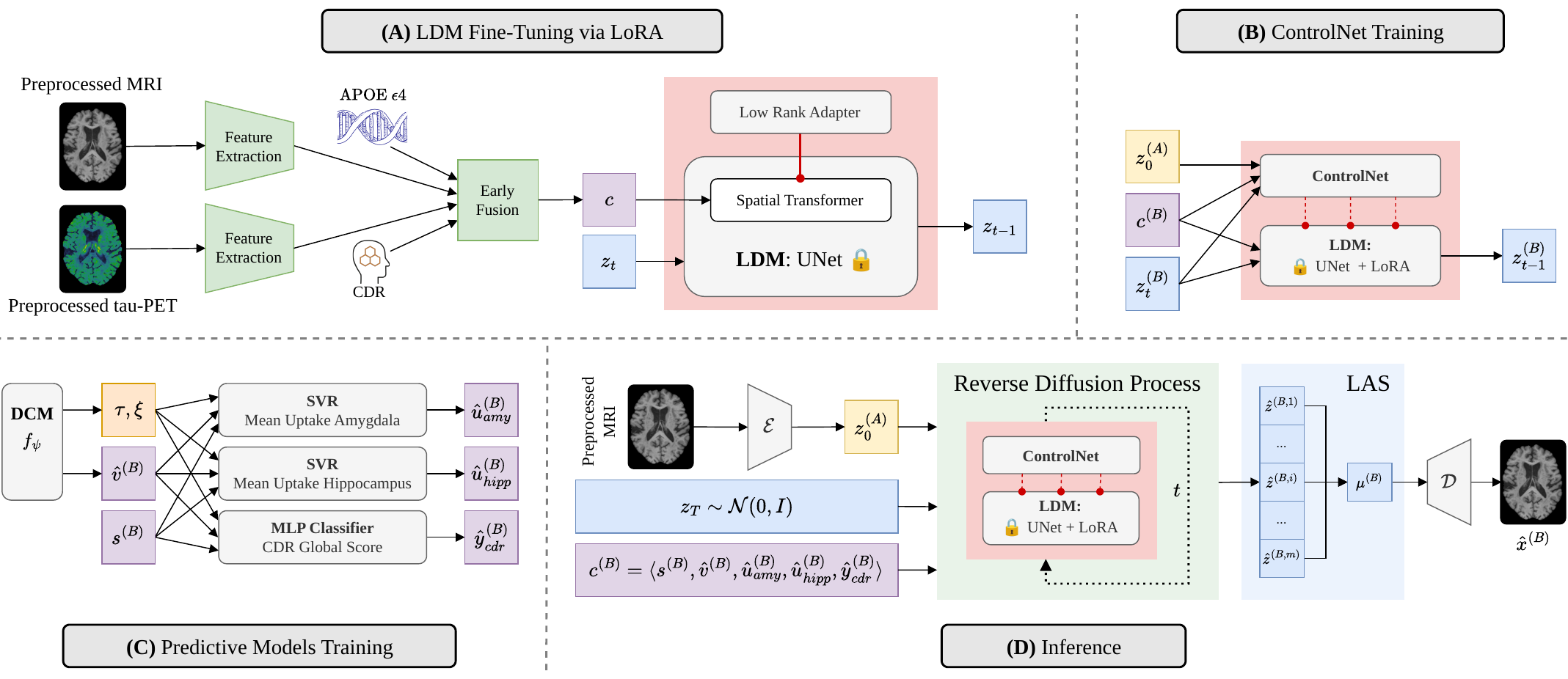}
    \caption{\label{fig:pipeline} Overview of the MC-LDM pipeline in four stages: (A) LoRA-based LDM fine-tuning to incorporate multimodal imaging, clinical, and genetic signals. (B) Training of a ControlNet branch to preserve subject-specific anatomy during generation. (C) Training of auxiliary predictive models that estimate future regional tau uptake and the CDR global score from MRI-derived features and covariates. (D) Inference stage, where the pretrained components are combined to synthesize future subject-specific MRIs. }
\end{figure*}

\subsection{Diffusion Models}

Denoising Diffusion Probabilistic Models (DDPMs)~\cite{Ddpm2020} learn to model the data distribution $p(x)$ by gradually reversing a predefined forward noising process. In the forward process, Gaussian noise is incrementally added to a data sample $x_0 \sim p_{\text{data}}(x)$ over $T$ steps according to a fixed Markov chain. For sufficiently large $T$, the resulting variable $x_T$ converges to pure Gaussian noise, $x_T \sim \mathcal{N}(0, I)$. Let $c$ denote a conditioning vector; the model then learns a denoising network $\epsilon_\theta(x_t, t, c)$ to approximate the reverse transition $p_\theta(x_{t-1} \mid x_t, c)$.

Latent Diffusion Models (LDMs)~\cite{Ldm2022} extend this paradigm by performing the diffusion process in a low-dimensional latent space $z = \mathcal{E}(x)$ obtained via an autoencoder, rather than in the high-dimensional image space. This enables more efficient training and high-resolution synthesis. While DDPMs operate directly in pixel space, LDMs shift the diffusion process into the latent domain. The reverse process operates on latent variables $z_t$: starting from Gaussian noise $z_T \sim \mathcal{N}(0, I)$, the model iteratively denoises the representation until $z_0$ is obtained, which is finally decoded as $\hat{x} = \mathcal{D}(z_0)$. As defined in~\cite{Ddpm2020}, the model is trained using the denoising score-matching objective:

\begin{equation}
    \mathcal{L}_{\epsilon} := \mathbb{E}_{t, x_0, \epsilon \sim \mathcal{N}(0,I)}
    \left[ \| \epsilon - \epsilon_\theta(z_t, t, c) \|_2^2 \right].
\end{equation}

\subsubsection{Conditional Generation via Cross-Attention}

While LDMs provide a powerful engine for image synthesis, targeted disease forecasting requires guiding this generative process with patient-specific multimodal priors. A common mechanism to achieve this is cross-attention, which is particularly well-suited for multimodal conditioning. A domain-specific encoder first maps the conditioning input $c$ into an intermediate representation $\tau_\theta(c)$. The U-Net~\cite{Unet2015} backbone is augmented with cross-attention layers, where spatial features of the noisy latent $z_t$ serve as queries ($Q$), while the projected condition $\tau_\theta(c)$ provides keys ($K$) and values ($V$)~\cite{Attention2017}. This allows the model to selectively attend to relevant conditioning information during denoising.

\subsubsection{Parameter-Efficient Fine-Tuning}

Fully fine-tuning diffusion models to incorporate new conditioning signals typically requires large datasets and can lead to catastrophic forgetting. To mitigate this, LoRA~\cite{LoRA2022} introduces a parameter-efficient adaptation strategy that keeps the original weights fixed while learning low-rank updates. Specifically, LoRA freezes the pre-trained weight matrix $W_0$ and optimizes a decomposition $\Delta W = BA$, where $B$ and $A$ are trainable low-rank matrices.

%% ---

\section{Methods}

In this section, we present our pipeline, comprising four stages depicted in Figure~\ref{fig:pipeline}: (A) LoRA-based LDM fine-tuning, (B) ControlNet training, (C) predictive models training, and (D) inference. All modules are trained sequentially, with no end-to-end optimization.

\subsection{Multimodal Conditioned LDM}

Our framework is built upon BrLP~\cite{BrLP2025} by adapting its LDM to support multimodal-conditioned forecasting under realistic clinical constraints (step A in Figure~\ref{fig:pipeline}). In particular, each MRI $x$ is encoded into a latent representation $z = \mathcal{E}(x)$ using an autoencoder $(\mathcal{E}, \mathcal{D})$. The diffusion model is trained using the standard noise prediction objective, minimizing the Mean Squared Error (MSE) between the true noise $\epsilon$ and the predicted noise $\epsilon_\theta(z_t, t, c)$, where $z_t$ denotes the noisy latent at timestep $t$ and $c$ is the multimodal conditioning vector. During training, all components of the conditioning vector $c$---age, sex, diagnosis, regional volumes, tau uptake, APOE $\varepsilon4$ status, CDR global score, and Total Intracranial Volume (TIV)---are extracted from ground-truth acquisitions, ensuring supervision from real molecular and clinical measurements. Instead of retraining the LDM backbone from scratch, LoRA injects trainable low-rank matrices into the query $Q$, key $K$, and value $V$ of both self-attention and cross-attention layers.  

While the LDM captures global disease-related information through $c$, it provides limited control over subject-specific anatomy. Following BrLP, we therefore subsequently train a ControlNet~\cite{ControlNet2023} branch (step B in Figure~\ref{fig:pipeline}) conditioned on the latent encoding of the baseline MRI, $z^{(A)} = \mathcal{E}(x^{(A)})$. During this stage, the LoRA-adapted U-Net is kept fully frozen, while only the ControlNet parameters $\phi$ are optimized. The ControlNet operates alongside the frozen U-Net during denoising, yielding a unified noise predictor $\epsilon_{\theta,\phi}(z_t, t, c, z^{(A)})$ without violating the sequential training scheme. This branch anchors the generation process to the patient’s baseline anatomy, encouraging the forecasted scan to preserve subject-specific structural characteristics while modeling their future evolution.

\subsection{Auxiliary Predictive Models} \label{sec:aux}

To condition the LDM, our framework requires estimates of the patient’s future biomarkers at the target age $B$. However, AD induces region-specific structural changes at varying rates~\cite{PINI201625}, making a direct mapping from baseline MRI to future multimodal biomarkers challenging.

We therefore first employ a Disease Course Mapping (DCM) framework~\cite{Koval2021}. Given a subject’s historical regional volumes and age, the DCM, denoted as $f_\psi$, models continuous volumetric trajectories to forecast future regional volumes $\hat{v}^{(B)}$ at target age $B$. It also estimates two subject-specific parameters: the time-shift $\tau$ (measured in years), which quantifies how much later or earlier the disease onset is compared to the population average; and the log-acceleration factor $\xi$, a scalar modulating the overall rate of volumetric decline (with positive values indicating faster progression than the average). We then compute the biological age $\Delta_{\text{bio}} = \text{age}_B - \tau$, which explicitly aligns the subject along a common pathological timeline.

We next define a target demographic vector $s^{(B)}$ containing $\text{age}_B$, $\text{diagnosis}_A$, sex, TIV, and APOE $\varepsilon4$ status. To strictly prevent temporal data leakage, all patient-specific variables are derived exclusively from the baseline visit. TIV is assumed constant throughout adulthood, while $\text{age}_B$ is specified by the user as the target forecasting age. We then concatenate $s^{(B)}$, $\Delta_{\text{bio}}$, and $\smash{\hat{v}^{(B)}}$ to train three independent auxiliary models responsible for estimating the future clinical and pathological descriptors required for the diffusion conditioning (step C in Figure~\ref{fig:pipeline}). Specifically, future mean tau uptake in the amygdala ($\smash{\hat{u}_\mathrm{amy}^{(B)}}$) and hippocampus ($\smash{\hat{u}_\mathrm{hipp}^{(B)}}$) are predicted via two separate RBF-kernel Support Vector Regressors (SVRs), while a Multi-Layer Perceptron (MLP) classifies the future CDR global score ($\smash{\hat{y}_\mathrm{cdr}^{(B)}}$).

These predictors are not intended as clinical surrogates for actual PET scans or formal cognitive assessments. Instead, their outputs act as informative priors that provide complementary multimodal guidance during forecasting.

\subsection{Inference Process}

Let $x^{(A)}$ denote the MRI acquired at age $A$, with its regional volumes $v^{(A)}$. Our goal is to synthesize the future MRI $\hat{x}^{(B)}$ at target age $B > A$. Because patient trajectories may be influenced by additional factors not explicitly modeled in our framework, including undocumented treatments, lifestyle variations, and comorbidities, our inference process represents future brain anatomy as a probability distribution over plausible outcomes rather than as a single deterministic prediction.

Since the final output must be a single MRI, we follow prior work that uses the average latent prediction as the final estimate, as this has been shown to produce robust results. More specifically, we adopt the Latent Average Stabilization (LAS) strategy introduced in~\cite{BrLP2025,Sargood2026}. Here, individual latent forecasts $\hat{z}^{(B,i)}$ can be viewed as samples from the model’s predictive distribution at age $B$. Averaging $m$ independently generated forecasts attenuates high-frequency stochastic variation and yields a more stable estimate of the expected latent representation, defined as:

\begin{equation}
\label{eq:las}
    \mu^{(B)} = \frac{1}{m} \sum_{i=1}^m \hat{z}^{(B,i)}.
\end{equation}

To operationally achieve this, the complete inference process (step D in Figure~\ref{fig:pipeline}) proceeds in five steps: \textbf{(i)} the DCM predicts future regional volumes and subject-specific progression parameters; \textbf{(ii)} the auxiliary models estimate the future tau uptakes and CDR score, which are concatenated with baseline patient metadata, the target age, and the structural predictions to form the final conditioning vector $c^{(B)}$; \textbf{(iii)} the baseline MRI is encoded into the latent representation $z^{(A)} = \mathcal{E}(x^{(A)})$; \textbf{(iv)} we generate the set of $m$ independent latent forecasts $\{\hat{z}^{(B,i)}\}_{i=1}^m$ by running the reverse diffusion process from distinct random noise seeds; and \textbf{(v)} we compute the expected representation $\mu^{(B)}$, which is finally decoded to obtain the forecasted image $\hat{x}^{(B)} = \mathcal{D}(\mu^{(B)})$.

\begin{table*}[t]
\small
\centering
\setlength{\tabcolsep}{7pt} 
\renewcommand{\arraystretch}{1.05}
\begin{tabular}{ l c c c c c }
\toprule
\multirow{2}{*}{\textbf{Setting}} 
& \multicolumn{2}{c}{\textcolor{teal}{\textbf{Image Metrics}}}
& \multicolumn{3}{c}{\textcolor{orange}{\textbf{Region Volumes MAE (\%)} $\downarrow$}} \\
& \textbf{SSIM} $\uparrow$ & \textbf{MSE} $\downarrow$
& \textbf{Hippocampus} & \textbf{Amygdala} & \textbf{Lat. Ventricles} \\

\midrule
\multicolumn{6}{c}{(A) Integration Strategy Ablation} \\
Direct Fusion & 0.922 $\pm$ 0.014 & 0.004 $\pm$ 0.001 & 0.049 $\pm$ 0.037 & 0.019 $\pm$ 0.016 & 0.501 $\pm$ 0.342 \\
\best{LoRA-based} & \best{0.931 $\pm$ 0.011} & \best{0.003 $\pm$ 0.001} & \best{0.034 $\pm$ 0.029} & \best{0.016 $\pm$ 0.011} & \best{0.263 $\pm$ 0.260} \\

\midrule
\multicolumn{6}{c}{(B) Conditioning Modalities Ablation} \\
Tau-PET Removed & 0.929 $\pm$ 0.015 & 0.004 $\pm$ 0.001 & 0.038 $\pm$ 0.034 & 0.019 $\pm$ 0.017 & 0.336 $\pm$ 0.474 \\
CDR Removed & 0.929 $\pm$ 0.016 & 0.004 $\pm$ 0.001 & 0.038 $\pm$ 0.035 & 0.019 $\pm$ 0.017 & 0.330 $\pm$ 0.489 \\
APOE $\varepsilon4$ Removed & 0.929 $\pm$ 0.016 & 0.004 $\pm$ 0.001 & 0.037 $\pm$ 0.034 & 0.018 $\pm$ 0.016 & 0.323 $\pm$ 0.471 \\
\best{All Modalities} & \best{0.931 $\pm$ 0.011} & \best{0.003 $\pm$ 0.001} & \best{0.034 $\pm$ 0.029} & \best{0.016 $\pm$ 0.011} & \best{0.263 $\pm$ 0.260} \\

\midrule
\multicolumn{6}{c}{(C) Sensitivity to Predicted Biomarkers} \\
Predicted Biomarkers & 0.931 $\pm$ 0.011 & 0.003 $\pm$ 0.001 & 0.034 $\pm$ 0.029 & 0.016 $\pm$ 0.011 & 0.263 $\pm$ 0.260 \\
\best{Oracle Biomarkers} & \best{0.932 $\pm$ 0.011} & \best{0.003 $\pm$ 0.001} & \best{0.028 $\pm$ 0.025} & \best{0.015 $\pm$ 0.010} & \best{0.258 $\pm$ 0.264} \\

\bottomrule
\end{tabular}
\caption{ \label{tab:ablation} \textbf{Ablation Study.} MRI forecasting results for the ablation studies, shown as SSIM, MSE, and region-wise volumetric MAE (\%) for AD-relevant structures. Values are reported as mean $\pm$ SD. Best results are highlighted in bold with a gray background.}
\end{table*}

\subsection{Pipeline Settings}

We use the U-Net and ControlNet implementations provided by the MONAI Generative Models library~\cite{pinaya2023,cardoso2022}. All the auxiliary predictive models were implemented using the \texttt{scikit-learn} library~\cite{Pedregosa2011}. The autoencoder and the DCM are adopted from BrLP~\cite{BrLP2025}, using their publicly available pre-trained model.

The pre-trained LDM backbone is kept frozen during training. We inject LoRA adapters into the U-Net, which are optimized for 120 epochs using the AdamW~\cite{loshchilov2018} optimizer with a learning rate of $5 \times 10^{-4}$, a weight decay of 0.01, and a batch size of 8. LoRA modules use rank $r = 32$, scaling factor $\alpha = 64$, and dropout $p = 0.1$. The ControlNet parameters $\phi$ are randomly initialized and trained under the same settings as in BrLP~\cite{BrLP2025}. At inference, we set the LAS hyperparameter $m = 8$.

All model training procedures and experiments were performed on a workstation equipped with an \textsc{Intel(R) Core(TM) i7-14700} CPU, $32$ GB of RAM, and a single \textsc{NVIDIA RTX 4070 Ti SUPER GPU} (16 GB of VRAM).

%% ---

\section{Experiments and Results}

This section evaluates the proposed MC-LDM framework. We first describe the clinical dataset and evaluation criteria in detail, followed by ablation studies to validate our architectural choices and multimodal integration strategy. Finally, we benchmark our approach against state-of-the-art methods through quantitative and qualitative analyses.

\subsection{Dataset}

The dataset includes 1,355 visits corresponding to 852 subjects from the Alzheimer’s Disease Neuroimaging Initiative (ADNI) 1/2/3/4/GO cohorts~\cite{Adni}. The cohort has a mean age of $73.0 \pm 8.0$ years and comprises 55.7\% female and 44.3\% male patients. Each visit includes a T1-weighted brain MRI, a tau-PET scan acquired with the $^{18}$F-AV1451 tracer, a CDR global score, and APOE $\varepsilon4$ status. Following the protocol described in~\cite{Scardace2026}, MRI and tau-PET scans were grouped into the same multimodal visit if acquired within a 6-month window~\cite{Scans6Months}, while the CDR global score was required to fall within a 12-month window~\cite{Cdr12Months}. All MRI and tau-PET scans were then preprocessed using the \texttt{neuroflux} pipeline~\cite{Scardace2026, scardace2026neuroflux}.

To rigorously prevent data leakage, all dataset partitions were strictly defined at the subject level using unique study identifiers (Roster IDs), ensuring that all longitudinal visits from a single individual remain within the same partition. For model development, the 1,355 visits were partitioned into training and validation sets using a 75/25 subject-level split, resulting in 639 training subjects (1,080 visits) and 213 validation subjects (275 visits). An independent test set comprising 664 visits from 197 subjects within the same ADNI cohorts was used for evaluation and includes only T1-weighted MRI and APOE $\varepsilon4$ status. Thus, the complete study encompasses 1,049 unique subjects, with zero overlap between the development and test sets.

\subsection{Auxiliary Predictive Models Evaluation}

To evaluate the auxiliary predictors described in Section~\ref{sec:aux}, we assess their standalone performance using 5-fold nested cross-validation and fit the final models exclusively on the LDM training set. The SVR predictors achieved a Mean Absolute Error (MAE) of $0.220 \pm 0.018$ for the amygdala and $0.168 \pm 0.013$ for the hippocampus. These errors correspond to 8.2\% and 7.0\% of their respective observed ranges ($0.30$–$2.99$ and $0.37$–$2.77$). All tau uptake measurements, including errors and ranges, are reported as Standardized Uptake Value Ratios. The CDR global score classifier achieved an accuracy of $82.03\%$. 

Overall, these results indicate that structural MRI and basic demographic features provide predictive information about molecular and cognitive states. The auxiliary predictors therefore provide informative priors for the MC-LDM generator, enabling it to operate without future tau-PET and clinical assessments at inference time.

\subsection{Evaluation Metrics}

We evaluate MC-LDM using \textbf{(i)} image-level metrics---Structural Similarity Index (SSIM) and MSE---which capture voxel-wise differences and structural consistency, and \textbf{(ii)} region-wise volumetric errors, assessed via MAE with respect to true follow-up scans. We focus on three AD-relevant structures---the hippocampus, amygdala, and lateral ventricles---known to exhibit characteristic atrophy and ventricular enlargement during disease progression.

\begin{table*}[t]
\small
\centering
\setlength{\tabcolsep}{7pt} 
\renewcommand{\arraystretch}{1.05}
\begin{tabular}{ l c c c c c }
\toprule

\multirow{2}{*}{\textbf{Method}} 
& \multicolumn{2}{c}{\textcolor{teal}{\textbf{Image Metrics}}}
& \multicolumn{3}{c}{\textcolor{orange}{\textbf{Region Volumes MAE (\%)} $\downarrow$}} \\
& \textbf{SSIM} $\uparrow$ & \textbf{MSE} $\downarrow$
& \textbf{Hippocampus} & \textbf{Amygdala} & \textbf{Lat. Ventricles} \\
    
\midrule
\multicolumn{6}{c}{(A) Cross-Sectional} \\

CounterSynth~\cite{CounterSynth2023} & 0.825 $\pm$ 0.071 & 0.013 $\pm$ 0.010 & 0.053 $\pm$ 0.028 & 0.019 $\pm$ 0.012 & 0.509 $\pm$ 0.460 \\
BrLP~\cite{BrLP2025} & 0.915 $\pm$ 0.038 & 0.005 $\pm$ 0.005 & 0.039 $\pm$ 0.033 & 0.020 $\pm$ 0.016 & 0.407 $\pm$ 0.444 \\
TADM-3D~\cite{Tadm3d2026} & 0.573 $\pm$ 0.019 & 0.134 $\pm$ 0.010 & 0.045 $\pm$ 0.043 & 0.021 $\pm$ 0.021 & 0.417 $\pm$ 0.515  \\
\best{MC-LDM (Proposed)} & \best{0.924 $\pm$ 0.036}$^*$ & \best{0.004 $\pm$ 0.005}$^*$ & \best{0.036 $\pm$ 0.029}$^*$ & \best{0.016 $\pm$ 0.012}$^*$ & \best{0.291 $\pm$ 0.279}$^*$ \\

\midrule
\multicolumn{6}{c}{(B) Longitudinal} \\
BrLP~\cite{BrLP2025} & 0.917 $\pm$ 0.035 & 0.005 $\pm$ 0.004 & 0.035 $\pm$ 0.031 & 0.019 $\pm$ 0.015 & 0.332 $\pm$ 0.325 \\
\best{MC-LDM (Proposed)} & \best{0.926 $\pm$ 0.029}$^*$ & \best{0.004 $\pm$ 0.003}$^*$ & \best{0.035 $\pm$ 0.028} & \best{0.016 $\pm$ 0.012}$^*$ & \best{0.265 $\pm$ 0.237}$^*$ \\

\bottomrule
\end{tabular}
\caption{ \label{tab:results} \textbf{Quantitative Comparison on ADNI.} Cross-sectional and longitudinal MRI forecasting results, shown as SSIM, MSE, and region-wise volumetric MAE (\%) for AD-relevant structures (hippocampus, amygdala, lateral ventricles). Values are reported as mean $\pm$ SD. Best results are highlighted in bold with a gray background. Asterisks denote a statistically significant improvement over the second-best performing method ($p < 0.05$, evaluated via paired $t$-tests). }
\end{table*}

\subsection{Ablation Study}

To evaluate our design choices and identify the main contributors to model performance, we conducted an ablation analysis on the validation set covering three aspects: the integration strategy for multimodal conditioning, the contribution of individual modalities, and the impact of errors in the predicted biomarkers. Statistical significance was assessed using paired $t$-tests with Benjamini--Hochberg correction for multiple comparisons, while effect sizes were quantified using paired Cohen’s $d$.

\subsubsection{Multimodal Integration Strategy}

We first compared two strategies for integrating multimodal covariates into the pre-trained LDM: direct fusion, where the new multimodal features are concatenated with the existing conditioning features and only the relevant cross-attention blocks are fine-tuned (25M trainable parameters), and the proposed LoRA-based modulation, which requires only 3M trainable parameters. As reported in Table~\ref{tab:ablation} (A), direct fusion leads to performance degradation, whereas LoRA-based integration provides a more effective adaptation of the pre-trained model. Specifically, it significantly reduces volumetric MAE by 30.6\% in the hippocampus, 15.8\% in the amygdala, and 47.5\% in the lateral ventricles, while reducing MSE by 25\% ($p < 0.05$, with large effect sizes $d > 1.0$ across all metrics), supporting LoRA as an effective strategy for multimodal conditioning.

\subsubsection{Contribution of the Clinical Modalities}

Using the LoRA-based configuration, we next evaluated the contribution of each clinical prior through a leave-one-modality-out analysis. As shown in Table~\ref{tab:ablation} (B), removing any modality degrades performance relative to the full model, although the magnitude of the effect varies across regions and metrics. Removing tau-PET produces the largest degradation in volumetric accuracy, increasing MAE by 27.7\% in the lateral ventricles, 18.7\% in the amygdala, and 11.8\% in the hippocampus. Removing CDR increases MSE by 33.3\% and lateral ventricles MAE by 25.4\%, while removing APOE $\varepsilon4$ increases lateral ventricles MAE by 22.8\% and hippocampus MAE by 8.8\%. Although these single-modality removals do not reach statistical significance ($p > 0.05$), the consistent degradation across anatomical regions and error metrics suggests that molecular, cognitive, and genetic priors provide complementary prognostic information beyond baseline anatomy alone.

\subsubsection{Predicted Biomarker Error Propagation}

Finally, we assessed the effect of errors introduced by the auxiliary predictors in our pipeline. We compared the fully automated pipeline with an oracle configuration conditioned on the ground-truth future biomarkers, thereby isolating the impact of auxiliary prediction errors. As shown in Table~\ref{tab:ablation} (C), the predicted-biomarker configuration produces only small absolute increases in volumetric MAE, of 1.9\% in the lateral ventricles and 6.6\% in the amygdala, while SSIM decreases by only 0.1\%. No statistically significant differences were observed across the evaluated metrics, indicating that forecasting performance is relatively insensitive to auxiliary prediction errors and supporting the use of predicted biomarkers when future multimodal assessments are unavailable at inference.

\subsection{Benchmarks}

\begin{figure*}[t]
    \centering
    \includegraphics[width=0.95\textwidth]{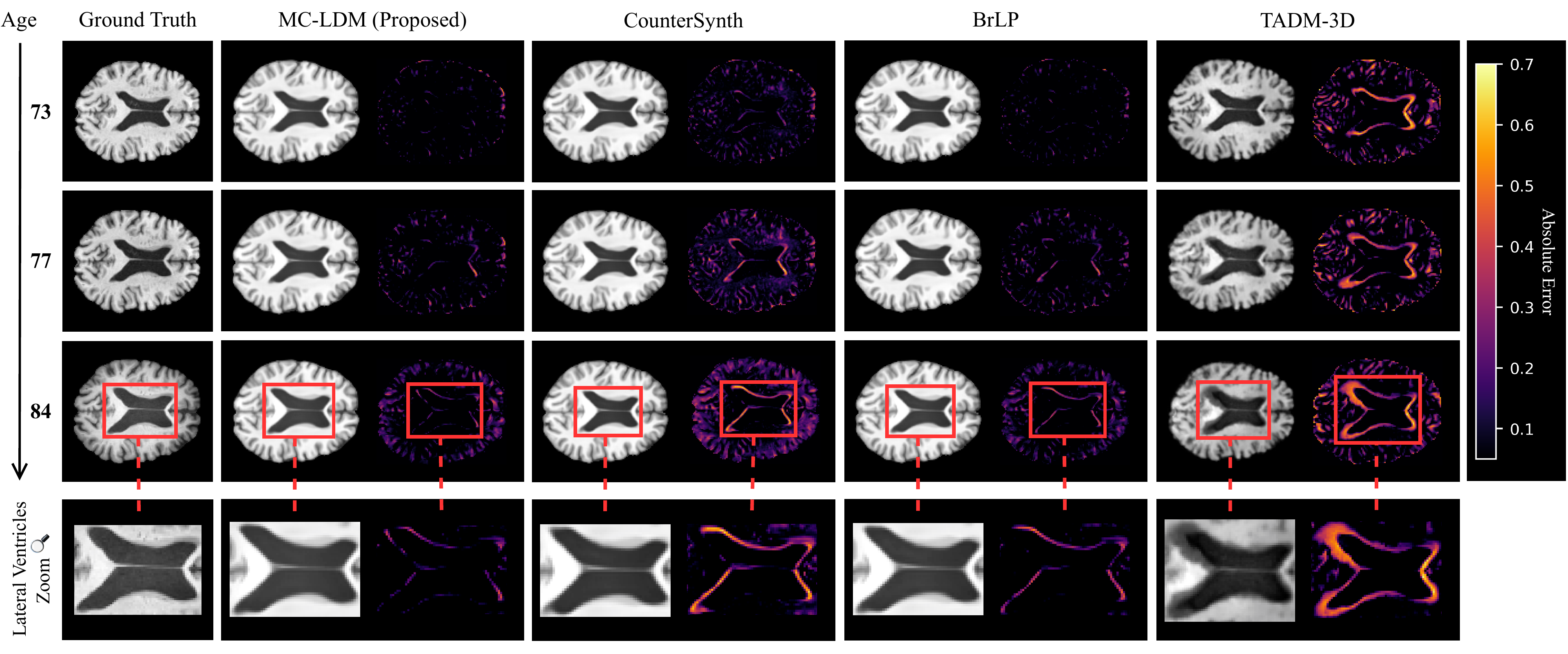}
    \caption{\label{fig:diffmap} \textbf{Qualitative Comparison.} Temporal progression on a 70-year-old subject from the test set, generated by our proposed approach against CounterSynth~\cite{CounterSynth2023}, BrLP~\cite{BrLP2025}, and TADM-3D~\cite{Tadm3d2026}. The figure illustrates predicted MRIs on the central axial slice (left side) and the corresponding absolute error maps (right side), highlighting regions of morphological deviation. The bottom row provides a magnified view of the lateral ventricles, with the selected region denoted by the bounding box in the original scans. }
\end{figure*}

We compare our framework against three spatiotemporal generative models: CounterSynth~\cite{CounterSynth2023}, TADM-3D~\cite{Tadm3d2026}, and BrLP~\cite{BrLP2025}. In particular, BrLP serves as our direct MRI-only baseline to quantify the performance gains obtained from multimodal conditioning. All baselines were implemented using their official, publicly available code and evaluated under the same data and evaluation procedures, ensuring a fair and consistent comparison.

We evaluate these methods across two settings: cross-sectional forecasting, which relies on a single baseline MRI, and longitudinal forecasting, which exploits multiple historical scans when available.

\subsubsection{Quantitative Results}

As shown in Table~\ref{tab:results}, MC-LDM achieves the best overall performance. In the cross-sectional setting, it outperforms the second-best method (BrLP) across all image-level metrics, improving SSIM by 0.9\% and reducing MSE by 20\%, indicating better structural fidelity. Regarding volumetric accuracy in AD-related regions, MC-LDM consistently achieves lower MAE, with error reductions of approximately 7.7\% in the hippocampus, 20\% in the amygdala, and 28.5\% in the lateral ventricles. In the longitudinal setting, similar improvements are observed, with SSIM increasing by approximately 1.0\% and MSE decreasing by 20\%. Volumetric MAE is further reduced by 15.8\% in the amygdala and 20.2\% in the lateral ventricles.

\subsubsection{Qualitative Results}

Finally, Figure~\ref{fig:diffmap} provides a qualitative comparison of our approach against other methods on a 70-year-old subject followed over 14 years. CounterSynth and TADM-3D exhibit noticeable spatiotemporal inconsistencies and fail to capture key structural changes observed in the ground-truth progression. While BrLP produces more coherent forecasts, our approach yields predictions that more closely align with the ground-truth evolution.

By consistently outperforming the MRI-only baseline (BrLP), our framework supports its core hypothesis: enriching the generative process with molecular and cognitive priors enables more accurate modeling of heterogeneous AD progression than structural anatomy alone.

\section{Conclusions}

We introduced MC-LDM, a multimodal-conditioned latent diffusion framework for patient-specific brain MRI forecasting that integrates heterogeneous disease descriptors within a single generative model. By leveraging rich multimodal supervision during training, MC-LDM remains applicable at inference using only baseline MRI, APOE $\varepsilon4$, and available patient metadata, enabling disease progression modeling under limited longitudinal data. High-value modalities such as tau-PET provide direct pathological insight but are costly and sparsely acquired, limiting the applicability of conventional multimodal methods. MC-LDM learns statistical relationships between molecular, clinical, and genetic factors and future anatomical changes, while replacing unavailable modalities at inference with auxiliary predictions. Beyond quantitative improvements, our work provides a step toward multimodal DPM for patient-specific 3D brain forecasting under realistic clinical constraints through an asymmetric conditioning paradigm that decouples multimodal supervision during training from modality availability at deployment.

Despite these promising results, some limitations remain. Our framework relies on a sequential training strategy, which may lead to error accumulation across intermediate stages. In addition, our evaluation is restricted to the ADNI cohort. Although external evaluation is important to assess generalizability, it is currently constrained by the limited availability of independent longitudinal cohorts with the specific combination of multimodal acquisitions, such as concurrent MRI and tau-PET, required by our pipeline. As more comprehensive longitudinal datasets become available, future work will prioritize rigorous cross-cohort evaluation. Further ablation studies will also investigate the individual contributions of multimodal conditioning and parameter-efficient fine-tuning, including a comparison with a same-architecture BrLP baseline using LoRA fine-tuning without multimodal conditioning, to better clarify the sources of the observed improvements.

%% ---

\section*{Acknowledgments}

D.R. was partially funded by the Italian Ministry of University and Research (MUR) under the Fondo Italiano per la Scienza (FIS 3), project “AI-Powered Digital Twins for Neurodegeneration Research: Advancing Disease Progression Modeling (NeuroTwin)” (CUP J53C25002200001).

\section*{Conflicts of Interest}

S.I. reports payment or honoraria and having stock or stock options from \textsc{Cerebriu A/S}.

%% ---

{
    \small
    \bibliographystyle{ieeenat_fullname}
    \bibliography{main}
}

\end{document}